\documentclass[runningheads]{llncs}
\usepackage{graphicx}
\usepackage{hyperref}
\usepackage{color}

\begin{document}
\title{The PV-ALE Dataset: Enhancing Apple Leaf Disease Classification Through Transfer Learning with Convolutional Neural Networks}
\titlerunning{The PV-ALE Dataset}
%
\author{Joseph Damilola Akinyemi\inst{1}\orcidID{0000-0003-3121-4231} \and
Kolawole John Adebayo\inst{2}\orcidID{0000-0001-7126-7026} } 
%
\authorrunning{J.D. Akinyemi and K.J. Adebayo}
%
\institute{University of York, York, United Kingdom \and
Dublin City University, Ireland\\
\email{joseph.akinyemi@york.ac.uk}\\
\email{kolawolejohn.adebayo@dcu.ie}}
%
\maketitle            
\begin{abstract}
As the global food security landscape continues to evolve, the need for accurate and reliable crop disease diagnosis has never been more pressing. To address global food security concerns, we extend the widely used PlantVillage dataset with additional apple leaf disease classes, enhancing diversity and complexity. Experimental evaluations on both original and extended datasets reveal that existing models struggle with the new additions, highlighting the need for more robust and generalizable computer vision models. Test F1-scores of 99.63\% and 97.87\% were obtained on the original and extended datasets, respectively. Our study provides a more challenging and diverse benchmark, paving the way for the development of accurate and reliable models for identifying apple leaf diseases under varying imaging conditions. The expanded dataset is available on \href{https://www.kaggle.com/datasets/akinyemijoseph/apple-leaf-disease-dataset-6-classes-v2}{Kaggle}, enabling future research to build upon our findings.

\end{abstract}
%

\section{Introduction}
\label{sec:intro}

Plant diseases pose a significant threat to plant life, crop yield, and food security, with far-reaching consequences for global food systems. The diverse nature of these diseases, which vary across plant species, necessitates specialized domain knowledge for effective detection and management. ~\cite{albattah2022novel}. Moreover, timely intervention to manage plant diseases and the need for efficient diagnostic methods have become increasingly pressing in the face of climate change, which continues to disrupt agricultural productivity. ~\cite{lineham2012food}. Conventionally, farmers utilise visual cues to identify infected crops. Even though a time-tested and efficient approach, it is time-consuming, labour-intensive and prone to inconsistent diagnoses due to human error or subjective interpretations, leading to the possibility of misclassifying unhealthy plants as healthy and vice versa ~\cite{jiang2019real}.

\noindent Apple trees are susceptible to a range of diseases including Powdery Mildew (PM), Apple Scab (AS), Alternaria leaf spot, Apple rust, etc., which can result in severe yield losses if left untreated ~\cite{ristaino2021persistent}. The timely identification of these diseases at their early stages is crucial for implementing effective control measures, which can mitigate the economic and environmental impacts of these diseases ~\cite{khan2022deep}. Recent advances in machine learning and deep learning have created new opportunities for automating plant disease detection, enabling the development of more accurate and efficient diagnostic tools ~\cite{thapa2020plant,li2021apple}.

\noindent We propose a deep learning-based approach for detecting apple leaf diseases from images, leveraging the economic importance of apple crops and the availability of high-quality, well-annotated open-source datasets. Our system aggregates 5 unique apple leaf disease types and 1 healthy class from consolidated existing and new datasets.

The primary contributions of our work include:
\begin{enumerate}
    \item A comprehensive consolidated dataset of apple leaf images annotated with corresponding disease labels.
    \item An efficient CNN architecture tailored specifically for the task of multi-class classification within the context of apple leaf disease detection.
    \item Rigorous evaluation metrics assessing model performance under various scenarios including class imbalance.
    \item Demonstration of superior accuracy compared to existing methods highlighting its potential application in real-world settings.
\end{enumerate}

The remaining parts of the paper are organized as follows: Sect. ~\ref{sec:lit_rev} presents a detailed review of the existing literature on apple leaf disease classification, Sect. ~\ref{sec:method} describes the dataset and method used in our work for apple leaf disease classification, Sect. ~\ref{sec:results} presents our experiments, results and comparative analysis and Sect. ~\ref{sec:conc} concludes the paper.

\section{Literature Review}
\label{sec:lit_rev}

Machine Learning has been used for plant disease detection with a reasonable degree of success in several studies ~\cite{tian2021diagnosis,albattah2022novel,si2024dual}. As in many other domains, Deep Learning algorithms have exceeded traditional Machine Learning algorithms in Apple disease detection and classification ~\cite{bansal2021disease}.

\noindent Jiang et al. ~\cite{jiang2019real} used an Inception network with Rainbow concatenation for Apple disease classification task on five classes of apple diseases (Alternaria leaf spot, Brown spot, Mosaic, Grey spot, and Rust). The experiments were conducted on 26,377 images of the Apple Leaf Disease Dataset (ALDD) dataset and reported a mean average accuracy of 78.8\% on a hold-out set of the ALDD. The authors of ~\cite{zhong2020research} employed a DenseNet-121 deep convolution network, formulating the problem as a multi-label classification task. The imbalanced nature of most apple disease datasets means that some disease classes will have very small probabilities, so the authors introduced a focus loss function instead of entropy loss, thereby increasing the performance of their model from 92.01\% to 93.71\%. In solving the class imbalance, Tian et al. ~\cite{tian2019detection} proposed using CycleGAN for data augmentation, thereby increasing the size and diversity of the dataset. Moreover, the authors employed DenseNet to optimize certain layers of the YOLO-V3 model. To conduct experiments, they collected 140 apple fruit images (increased to 700 images through augmentation) and obtained an accuracy of 95.57\%. 

\noindent Recent studies have proposed various deep learning-based approaches for plant disease detection. For instance, Tian et al. ~\cite{tian2021diagnosis} introduced a Multi-scale Dense classification network that achieved state-of-the-art classification accuracies of 94.31\% and 94.74\% on a dataset of 11 classes, including healthy and diseased apple fruits and leaves. The study employed Cycle-GAN for data augmentation to address the challenge of insufficient images. Similarly, the authors in ~\cite{albattah2022novel} proposed a three-stage approach for plant disease detection, achieving an accuracy of 99.98\% on the PlantVillage dataset ~\cite{thapa2020plant} using DenseNet ~\cite{huang2017densely}.

\noindent Other deep learning-based studies have mostly involved various network architectures on different datasets using different evaluation strategies, thus making it difficult to make direct comparisons among studies. For instance, the authors in ~\cite{khan2022deep} proposed a 2-stage approach using the Xception model to first extract low-level features and then Faster-RCNN to localize the diseased image region. However, their method achieved 88\% accuracy which seems subpar compared to simpler CNN-based studies such as ~\cite{bansal2021disease,li2021apple}

\noindent The reviewed studies underscore the potential of deep learning algorithms and related techniques in apple disease detection. However, there remains a need for further research to enhance the accuracy and robustness of these models. First, the existing apple leaf datasets, while valuable, present certain limitations that necessitate further research. On primary gap is the variability in disease classes across different datasets. Most datasets focus on a different set of disease classes, and there is a scarcity of comprehensive datasets encompassing a wide range of these different disease classes. Moreover, most of the existing datasets are not very large and the few large ones are not publicly available. These limitations restrict the generalizability of the models trained on these datasets, as they may not perform well when confronted with diseases outside those in their training set. 


\noindent This paper aims to fill these gaps by extending a well-used existing dataset of apple leaf diseases using validated manual data collection and augmentation techniques to create a larger and more comprehensive dataset. By doing so, we hope to enhance the generalizability and robustness of apple disease detection models. This research is thus a significant step towards harnessing the power of deep learning to address the challenges in apple disease detection, promoting sustainable agriculture.

\section{Methodology}
\label{sec:method}
\subsection{Dataset}

\noindent Despite the significant research efforts in tackling apple disease detection, the available datasets often present certain limitations that hinder comprehensive and robust model training. These datasets are typically small-sized, imbalanced, and contain a limited number of disease classes ~\cite{li2021apple}. Moreover, they are not often publicly available, primarily due to the substantial expert effort required for their collection and annotation.

\noindent One such dataset is PlantVillage ~\cite{thapa2020plant}, the most commonly used dataset in this domain. It contains only 3,171 images of apple leaves, more than half of which are healthy leaves. The remaining half is distributed among just three disease classes: \textit{rust}, \textit{black rot}, and \textit{scab}. The high-resolution, single-leaf images in the PlantVillage dataset, while useful, present less of a challenge and are less applicable in real-world situations, which often involve variable image quality, multiple leaves, and complex backgrounds.

\noindent To address these challenges, we collected apple leaf images for two additional disease classes: \textit{Alternaria leaf spot} and \textit{Powdery Mildew}. These images were collected via Internet image search using the disease names as keywords. Initially, we gathered 79 and 183 images for \textit{Alternaria leaf spot} and \textit{Powdery Mildew}, respectively. To enhance the data pre-processing and preparation steps, and ensure the quality and reliability of our dataset, we followed the guideline below:
\begin{enumerate}
    \item Image Download: Download only images containing apples and apple leaves. This ensures that our dataset is specific to our research focus and reduces the likelihood of including irrelevant images.
    \item Caption Verification: Verify the image caption to ensure it bears the disease name with respect to the query. This step helps to confirm that the image is indeed related to the disease class it is supposed to represent.
    \item URL Verification: Verify the image URL to ensure the caption is trustworthy. The URL should be from a publication or website content of an agricultural-focused research performing organization, laboratory or nursery. By doing so, we can increase the likelihood of obtaining accurate and reliable images.
    \item Image Matching: Implement additional verification to ensure that the image matches or looks like other images from the same class/pool. This step helps to maintain consistency within each disease class and reduce the risk of mislabeling.
    \item Human Validation: To further ensure the quality of our dataset, we also use human annotators to validate each image and its corresponding disease class to ensure their accuracy. This step adds an extra layer of verification and helps to identify and correct any potential errors or inconsistencies in our dataset. For instance, where some images bearing tags related to the disease class were not visually related to the disease class, those were removed.
    \item Image Cropping: We cropped images to obtain multiple single-leaf images where possible. This step increased the number of images, as some images with leaf clusters generated more single-leaf images. However, many images of leaf clusters were still retained to increase the complexity and diversity of the dataset.
    \item Background Removal: We cropped the images to remove the background, focusing on the apple leaves and their diseases.
\end{enumerate}

\begin{figure}
  \centering
   \includegraphics[width=0.8\textwidth]{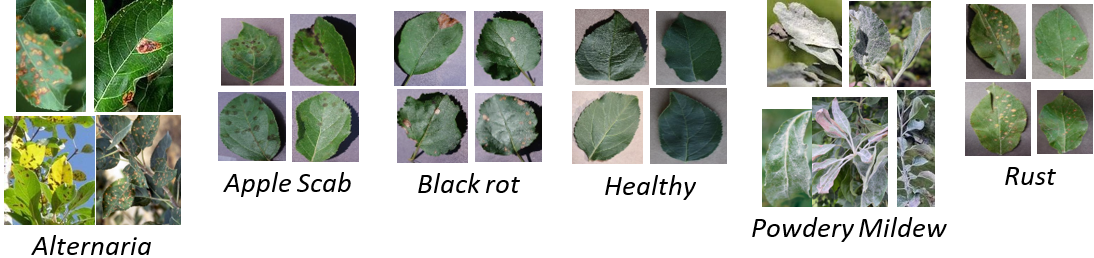}

   \caption{Sample images of apple leaves in the dataset.}
   \label{fig:fig1}
\end{figure}

\noindent We posit that adhering to this guideline significantly improves the quality, reliability, and diversity of our dataset, ultimately contributing to the development of more accurate and robust apple disease detection models. Fig.~\ref{fig:fig1} presents representative samples of images from the two newly introduced disease classes, namely (\textit{Alternaria} and \textit{Powdery Mildew}), alongside samples from the pre-existing classes in the PlantVillage dataset. Upon completion of the image collection and pre-processing stages, we obtained 85 images for \textit{Alternaria} and 127 images for \textit{Powdery Mildew}. As observed in Fig.~\ref{fig:fig1}, the newly incorporated classes introduce several challenges, such as variable-sized and small images, low resolution, the presence of both single-leaf and leaf-cluster images, complex backgrounds, and a limited number of images in each class. However, these characteristics contribute to a more challenging and realistic representation of real-world conditions, thereby enhancing the applicability and generalizability of our apple leaf disease detection models.

\noindent After the initial data curation process, we integrated the newly collected disease classes with the existing four classes of the PlantVillage dataset, resulting in a consolidated dataset of 3,383 images. To address the significant class imbalance, we divided the dataset into a 70\% training set and a 30\% testing set. We then applied various data augmentation techniques, such as angled image rotation, light intensity variation, and zooming, to balance the training set data. This augmentation process yielded 20,808 images fairly evenly distributed across the six classes, forming the PlantVillage Apple Leaves Extended (PV-ALE) dataset. PV-ALE can be accessed on \href{https://www.kaggle.com/datasets/akinyemijoseph/apple-leaf-disease-dataset-6-classes-v2}{Kaggle}.

\noindent The dataset statistics presented in Table~\ref{tab:table1} reveal a significant bias towards healthy leaves. Despite being significantly smaller, the two new classes constitute an essential addition of new disease classes to the entire dataset. Table~\ref{tab:table1} also demonstrates how the augmented images helped us achieve a fairly balanced distribution of instances per class, resulting in a training set with approximately 3,500 images in each class.

\noindent The PV-ALE dataset holds significance not only for the challenges it presents but also for expanding the existing classes in the most widely-used apple disease dataset, PlantVillage. Crucially, implementing class balancing and data augmentation methods generates a more diverse set of samples, satisfying the training and evaluation requirements for future research. Moreover, by open-sourcing the PV-ALE dataset, we aim to establish a more comprehensive benchmark with a broader range of classes and samples. This will facilitate more accurate comparisons among future studies, as many researchers have previously published experiments on various closed datasets, making it extremely difficult to objectively compare existing systems in the literature. In the subsequent sections, we offer a detailed account of our primary deep-learning classification models and the established baselines.

\begin{table}
\caption{Dataset statistics}
\centering
\begin{tabular}{|l|l|l|l|l|}
\hline
Classes & original set & training set & test set & augmented training set \\
\hline
Alternaria & 85 & 60 & 25 & 3420 \\
Apple Scab & 630 & 441 & 189 & 3528 \\
Black rot & 621 & 435 & 186 & 3480\\
Cedar Apple rust & 275 & 192 & 83 & 3456 \\
Healthy & 1645 & 1151 & 494 & 3453 \\
Powdery mildew & 127 & 89 & 38 & 3471 \\
Totals & 3383 & 2368 & 1015 & 20808 \\
\hline
\end{tabular}
\label{tab:table1}
\end{table}

\subsection{Apple disease classification}
The efficacy and utility of the proposed PV-ALE dataset were comprehensively evaluated through a series of experiments conducted on both the extended (PV-ALE) and the original (PV-AL) PlantVillage datasets. To assess the impact of the two newly incorporated classes on performance and to determine the potential contribution of the PV-ALE dataset in advancing research on apple leaf disease detection, we employed two distinct deep learning models for training and validation on each dataset, resulting in a total of four experiments.

\noindent The task of identifying or classifying apple leaf diseases based on their leaf images was formulated as a multiclass classification problem. Deep learning techniques were leveraged for both feature extraction and classification throughout the experiments. In one set of experiments, we adopted a transfer learning approach by fine-tuning a ResNet50 architecture ~\cite{he2016deep} pre-trained on the ImageNet dataset ~\cite{DenDon09Imagenet}. The ResNet model was utilized to extract salient features from the input leaf images, which were subsequently classified based on the extracted features. In the other set of experiments, we constructed a simple 7-layer Convolutional Neural Network (CNN) from scratch for the same purpose.

\noindent The selection of these models was motivated by the proven capability of CNNs and ResNet50 in extracting discriminative features from images, which can be highly descriptive and instrumental in distinguishing images based on subtle differences. In the context of this study, these subtle differences pertained to the specific disease types affecting the apple leaves.

\noindent In our transfer learning approach, we employed the \emph{ResNet50} model pre-trained on the ImageNet dataset as the backbone for feature extraction. The original classification layer was replaced with a custom head comprising a Global Average Pooling layer, a Flatten layer, and five Fully Connected layers interspersed with three Dropout layers, resulting in a total of over 24 million trainable parameters. The number of units in the Fully Connected (Dense) layers was empirically determined during hyperparameter tuning, with the final layer having six units corresponding to the six classes for the extended dataset or four units for the original four-class PlantVillage dataset. The Dropout layers randomly omitted either 30\% or 20\% of the neurons from the preceding layers. The model was trained using the Adam optimizer with a base learning rate of \emph{5e-5}, a categorical cross-entropy loss function, and a \emph{softmax} activation function for the final Dense layer, while rectified linear unit activations were employed for the remaining Dense layers.

\noindent Our custom 7-layer CNN architecture consisted of two \emph{2D} convolutional layers, a Flatten layer, and four Fully Connected (Dense) layers, complemented by intermittent pooling and Dropout layers to extract salient features and mitigate overfitting during training. With over 252 million trainable parameters, this model had significantly more parameters than our transfer learning approach, primarily due to the training of the entire CNN, including the convolutional layers with numerous \emph{3 × 3} filters, from scratch. The CNN was trained using the Adam optimizer with a \emph{5e-5} base learning rate, a categorical cross-entropy loss function, and rectified linear unit activation functions for all Dense layers except the final layer, which employed a Softmax activation function. Across all experiments, the input images were resized to \emph{225 × 225} pixels to conform to the ResNet input layer requirements.

\noindent To prevent overfitting and data leakage, we employed three strategies: a clear separation of training and test data, Dropout layers, and the Early Stopping technique. The test sets were strictly isolated from the training sets and models during training. For validation purposes, 10\% of the training set was held out as a validation set, allowing for fine-tuning and performance evaluation without exposing the models to the test set, thus preventing overfitting. This approach ensured a reliable assessment of the models' generalization capabilities on the test set. Additionally, Dropout layers were incorporated into each network architecture, randomly omitting \emph{20\%} to \emph{30\%} of the neurons from the preceding layer before propagating to the next layer. Furthermore, Early Stopping was implemented to monitor the validation loss and terminate training after 10 (for ResNet) or 5 (for CNN) consecutive epochs without any further reduction in validation loss. These measures collectively aimed to produce well-generalized models capable of robust performance on the test set. The difference in Early Stopping tolerance values between the ResNet and CNN models was to provide the ResNet with a higher chance of convergence and to reduce the CNN's training time, given its larger parameter count.

\section{Results}
\label{sec:results}

Previous works on apple leaf disease identification or classification have often reported high accuracy rates due to the low variability present in the datasets. Most datasets contain high-resolution single-leaf images captured against plain backgrounds. While this simplifies image processing, it fails to represent real-world scenarios where individual leaves are seldom examined against a plain background to determine plant diseases, especially in large-scale farming operations. The PV-ALE dataset collected for this work addresses this limitation by incorporating two new classes containing low-resolution images, leaf clusters, and complex backgrounds (often including other leaves or garden plants).

\noindent In this section, we report the results of our experiments using both the original PlantVillage dataset and the PV-ALE dataset. To ensure a fair comparison, the same network architectures and parameter settings were employed for both datasets. Additionally, we spot-checked the reported accuracies of some previous works on specific classes of apple leaf diseases included in our dataset and compared them with our results. All experiments were performed on Kaggle's cloud-based GPU scripting environment, utilizing a GPU P100 with 16GB GPU Memory and 32GB RAM.

\subsection{Results on the PV-ALE dataset} 
\label{sec:tl_pv-ale}

\begin{figure}
  \centering
   \includegraphics[width=\textwidth]{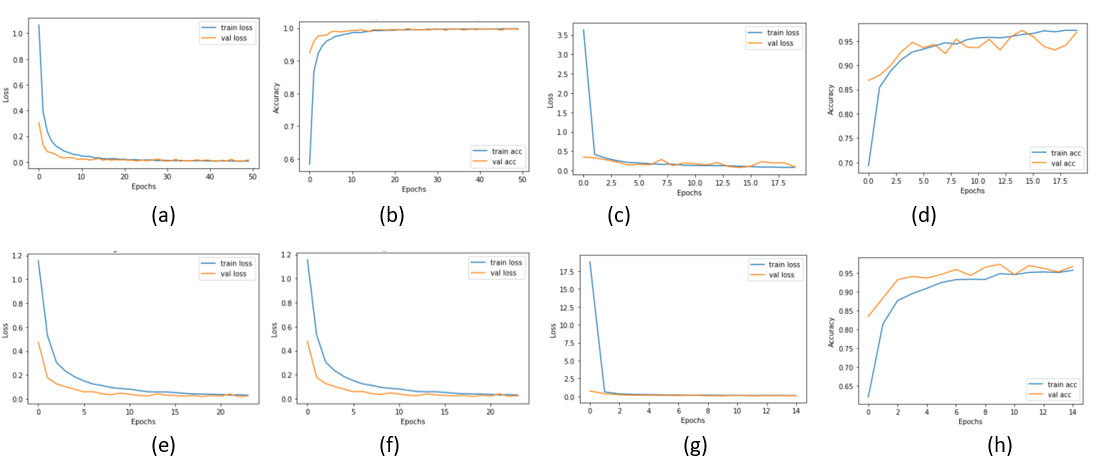}
   \caption{Loss and accuracy on PV-ALE (top row) and PV-AL (bottom row) datasets}
  \label{fig:figure2}
\end{figure}

\textbf{Transfer learning results on the PV-ALE dataset}.
The Transfer Learning (TL) method on the PlantVillage-ALE (PV-ALE) dataset demonstrated very promising results. The training and validation accuracy and loss curves for the ResNet50 model are shown in Fig.~\ref{fig:figure2}. Fig. 2a depicts the loss, while Fig. 2b illustrates the accuracy. Both figures indicate a smooth training progression with no evidence of overfitting, as the validation performance is at par with the training performance throughout the training process. It can be observed that from the very first epoch, both training and validation accuracy surpassed 90\%.  This is made possible by the depth of the ResNet architecture, which enables it to learn rich image features, and the added top layers effectively adapted those features to discriminate between the different diseases shown in the leaf images. Table~\ref{tab:table2} shows the results obtained by our TL method (ResNet50) on the PV-ALE and PV-AL datasets (rows 1 \& 3), which are impressive.

\begin{table}
\caption{Precision (Prec.), Recall (Rec.), F1-scores (F1) and Accuracy on PV-ALE and PV-AL datasets}
  \centering
  \begin{tabular}{|l|l|l|l|l|l|}
    \hline
    Dataset & Model & Precision (\%) & Recall (\%) & F1-score (\%) & Accuracy (\%)  \\
    \hline
    PV-ALE & ResNet50 & 99.06 & 96. 82 & 97.87 & 99.11 \\
    PV-ALE & CNN & 99.01 & 88.94 & 89.77 & 94.98 \\
    PV-AL & ResNet50 & 99.55 & 99.72 & 99.63 & 99.58 \\
    PV-AL & CNN & 94.37 & 95.82 & 95.05 & 95.59 \\
    \hline
  \end{tabular}
  \label{tab:table2}
\end{table}

\begin{figure}[t]
  \centering
   \includegraphics[width=1\linewidth]{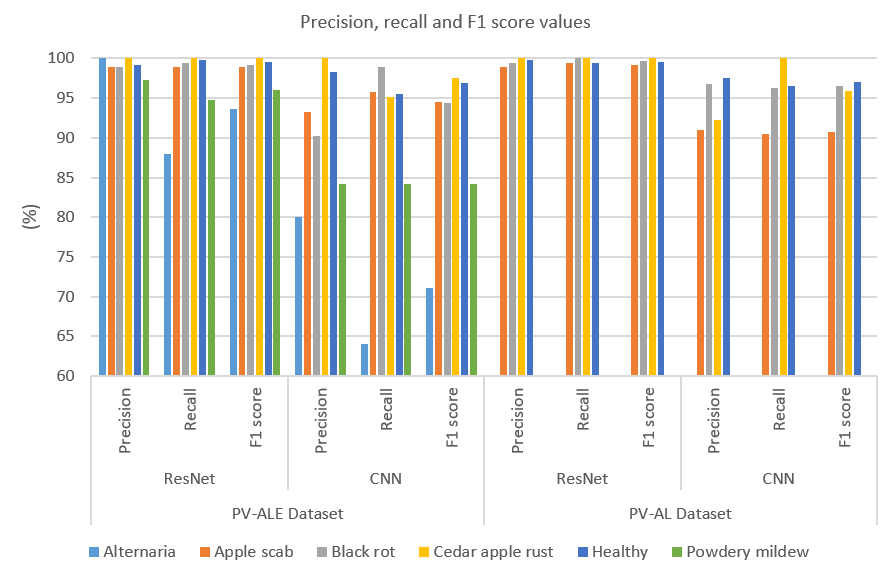}

   \caption{Precision, recall and F1-score values on each class of both datasets}
   \label{fig:figure3}
\end{figure}

\textbf{CNN results on the PV-ALE dataset}.
Similar to the trend observed with the TL technique, the simple CNN architecture we constructed also performed reasonably well on the training, validation and test datasets. As shown in figures Fig. 2c and Fig. 2d, the CNN training progressed well, though not as smoothly as the ResNet model, yet the validation progression generally matched the training progression. As presented in Table~\ref{tab:table2} (row 2), the CNN model achieves $\approx$95\% accuracy which is $\approx$4\% lower than that of the ResNet50 model. This small difference indicates that the CNN architecture is generally well-suited to the problem of identifying apple diseases from apple leaf images. 

\subsection{Results on the Original PV-AL Dataset}
\label{sec:result_pv-al}
The experiments conducted on the original PlantVillage Apple Leaf (PV-AL) dataset, comprising four classes (three disease classes and one healthy class), reveal that it is less challenging than the PV-ALE dataset. Both the ResNet and CNN models demonstrate this observation, as shown in Fig. ~\ref{fig:figure2} (bottom row). The validation accuracy and loss consistently outperform the training accuracy and loss for most of the training epochs. Despite an early stopping tolerance of 10 epochs, the ResNet model converges quickly after 24 epochs (Fig. 2e and 2f), which is only halfway through the number of epochs required for the same ResNet model to stop training on the PV-ALE dataset. Table~\ref{tab:table2} (rows 3 \& 4) shows that ResNet achieves an accuracy of 99.58\% and an F1 score of 99.63\% while CNN achieves an accuracy of 95.59\% and an F1 score of 95.05\% on the PV-AL dataset. The training progression of the CNN model indicates a good fit (Fig. 2g and 2h) as the validation performance surpasses the training performance throughout the training process. Without the Early Stopping setting, the CNN could potentially reach even higher accuracy in a few more epochs.

\subsection{Comparing results on both datasets}

Fig.~\ref{fig:figure3} illustrates the precision, recall and F1-score values obtained using the ResNet50 and CNN models on each class of both datasets. Generally, the \emph{Alternaria leaf spot} class (light blue bar) is the most challenging to identify while the \emph{Cedar Apple Rust} class (yellow bar) is almost perfectly identified in all cases. One possible explanation for this observation is the number of samples for each class in the test set; the \emph{Alternaria} class has the fewest samples, while the \emph{Cedar Apple Rust} class has the highest number of samples. However, this discrepancy in class distribution is an accurate reflection of the actual data and could not be circumvented. Notably, this result is not a function of the training data distribution, as the augmented training data is fairly balanced across all classes. Even though the \emph{Alternaria} class has the least number of samples (3420) in the augmented training set, the \emph{Cedar Apple Rust} class, which has the third lowest number of samples (3456), exhibits the best performance as shown in Fig.~\ref{fig:figure3}. This suggests that the presented results on the test set are more influenced by the visual cues responsible for each class rather than the class size. The network seems to recognize the \emph{Cedar Apple Rust} disease more easily than any other class. Interestingly, while one might expect the \emph{Healthy} class to be most easily recognised, this is not the case. This pattern is consistent throughout the experiments, with the \emph{Rust} class being the most accurately recognized, followed closely by the \textit{Healthy} and \textit{Black Rot} classes. This implies that the visual features on the \emph{cedar apple rust} leaves are easier for the networks to identify compared to other classes. 

Given that both datasets were subjected to the same network under identical parameter settings, the performance difference of approximately 2\% (F1 scores of 97.78\% on PV-ALE and 99.63\% on PV-AL, as shown in Table ~\ref{tab:table2} (rows 1 \& 3) and the fact that better performance was achieved on the PV-AL dataset in fewer training epochs indicate that the two new classes introduced additional complexities to the original dataset. These findings suggest that the PV-ALE dataset will prove resourceful for future research in this field. While one could argue that the PV-AL dataset is smaller than the PV-ALE dataset, the difference between the two test sets is only 63 samples, which is not up to the size of any of the two added classes and is about  6\% - 7\% of the total size of the test set for each dataset. These results reveal the need for more disease classes in the apple leaf disease dataset. 

\subsection{Analysis and Discussion}

Finally, we performed a comparative analysis of the classification accuracy for each class in the PV-ALE dataset with those reported in the literature. It is worth noting that most previous works often employed different disease classes, datasets, and dataset split ratios, making direct comparisons challenging. Therefore, our comparison is solely based on the reported accuracies for corresponding classes found in previous literature against those in our dataset. In cases where the number of class samples in the test set is included, we have stated this information as well. For each previous work, Table~\ref{tab:table3} reports the best F1 score for each class, as well as the overall F1 score across all classes, bearing in mind that the entire set of classes in the concerned datasets differs.

\begin{table}
\caption{Class-wise comparison of F1 scores (\%) of previous works on different apple disease datasets (number of instances per class in brackets).}
  \centering
  \begin{tabular}{|l|l|l|l|l|l|}
    \hline
     Work & \cite{zhong2020research} & \cite{li2021apple} & \cite{wang2021identification} & \cite{khan2022deep} & \textbf{Ours} \\   
     \hline
     Model & Dense-Net & RegNet & CA-ENet & Xcep. & ResNet \\
     Num. classes & 6 & 5 & 8 & 9 & 6 \\
     \hline
     Alternaria & - & - & - & 86 (116) & 88.9 (25) \\ 
     Scab & 73.7 (82) & 98.9 (46) & 99.8 (500) & 83 (212) & 96.9 (189) \\ 
     Rot & - & - & 98.9 (500) & - & 99.7 (186) \\ 
     Rust & 87.1 (42) & 99.1 (54) & 99.9 (500) & - & 100 (83) \\ 
     Powdery Mildew & - & - & - & 86 (85) & 92.1 (38) \\ 
     Healthy & 98.5 (127) & 98 (49) & 97.6 (500) & 97 (90) & 98.9 (494) \\ 
     All classes & 93.7 (493) & 99.2 (260) & 98.8 (4000) & 78.1 (686) & 97.9 (1015) \\
    
    \hline
  \end{tabular}
  \label{tab:table3}
\end{table}

From Table ~\ref{tab:table3}, one can observe the high variation in the class distributions as well as sample sizes across different works/datasets; making direct comparisons challenging. However, the table provides the following insights:
\begin{enumerate}
    \item The relative dataset sizes and number of samples in each class remain relatively small across most studies.
    \item Datasets containing larger sample sizes per class tend to achieve better overall performance, regardless of the number of classes.
    \item Our dataset remains significantly larger than most other datasets while maintaining approximately 50\% coverage of the classes present in other datasets.
    \item The two most well-predicted classes are Cedar Rust and Healthy leaves while Alternaria and Powdery Mildew are scarcely represented in most datasets.
\end{enumerate}

It is evident from the comparative analysis that the PV-ALE dataset poses a more challenging and realistic benchmark for apple leaf disease classification. While previous works have reported high accuracies on specific disease classes, their evaluations were often conducted on datasets with limited complexity. The inclusion of low-resolution images, leaf clusters, and complex backgrounds in the PV-ALE dataset introduces diversities that are more representative of the real world. Furthermore, the discrepancies in the reported accuracies across different classes highlight the importance of a comprehensive and diverse dataset. Certain disease classes, such as \emph{Cedar Apple Rust}, appear less challenging due to their distinct visual cues unlike \emph{Alternaria Leaf Spot} which has subtle visual manifestations or limited representation in the dataset.

These observations underscore the need for standardized and comprehensive datasets that encompass a wide range of disease types and imaging conditions. The PV-ALE dataset addresses this need by providing a diverse and challenging benchmark that can facilitate the development of robust and generalizable Computer Vision models for accurate apple leaf disease classification. It is important to note that while direct comparisons with previous works are difficult due to the aforementioned differences in datasets and experimental setups, the comparative analysis serves to highlight the advancements and challenges introduced by the PV-ALE dataset. 

\section{Conclusion}
\label{sec:conc}

This work presents PV-ALE, a more diverse and comprehensive Apple Disease dataset over the PlantVillage dataset ~\cite{thapa2020plant}. Extensive experiments were conducted using transfer learning with ResNet50 and a custom CNN model. The results demonstrated that while both models performed excellently on the two datasets, PV-ALE seemed more challenging. We conducted a class-wise comparison with previous works that employed various datasets, revealing a need for increased diversity in existing datasets. The PV-ALE dataset addresses this need by incorporating two new classes that are underrepresented in other datasets, thereby introducing much-needed diversity and complexity. A major limitation of this work is the relatively small size of the test set, which is a consequence of the size and distribution of the original dataset. In future works, we aim to further enhance the dataset by increasing the size of each class and covering a broader range of apple disease types.

\subsubsection{\ackname}
Kolawole Adebayo has been supported by Enterprise Ireland’s CareerFit-Plus Co-fund and the European Union’s Horizon 2020 research and innovation programme Marie Skłodowska-Curie Grant No. 847402.

%
%
\bibliographystyle{splncs04}
\bibliography{refs}

\end{document}